\documentclass{article}

\usepackage[preprint]{neurips_2025}

\usepackage[utf8]{inputenc} %
\usepackage[T1]{fontenc}    %
\usepackage{hyperref}       %
\usepackage{url}            %
\usepackage{booktabs}       %
\usepackage{amsfonts}       %
\usepackage{nicefrac}       %
\usepackage{microtype}      %
\usepackage{graphicx}
\usepackage{multirow}
\usepackage[table,xcdraw]{xcolor}
\usepackage{soul}
\usepackage{wrapfig}
\usepackage{float}
\usepackage{amsmath}

\title{SimpleGVR: A Simple Baseline for Latent-Cascaded Generative Video Super-Resolution}

\author{
    Liangbin Xie$^{1,2}$\footnotemark[1]  \hspace{9pt}  Yu Li$^{3}$ \hspace{9pt} Shian Du$^{3}$ \hspace{9pt} Menghan Xia$^{4}$\footnotemark[2] \hspace{9pt} Xintao Wang$^{4}$ \hspace{9pt} \\ 
    \textbf{Fanghua Yu$^{2}$ \hspace{9pt} Ziyan Chen$^{2}$ \hspace{9pt} Pengfei Wan$^{4}$ \hspace{9pt} Jiantao Zhou$^{1}$\footnotemark[2] \hspace{9pt} Chao Dong}$^{2,5}$ \\
	\small{$^{1}$State Key Laboratory of Internet of Things for Smart City, University of Macau} \\
	\small{$^{2}$Shenzhen Institutes of Advanced Technology, Chinese Academy of Sciences} \\
	\small{$^{3}$Tsinghua University} \hspace{9pt}
	\small{$^{4}$Kuaishou Technology} \hspace{9pt}
        \small{$^{5}$Shenzhen University of Advanced Technology} \\
}

\begin{document}

\maketitle
\let\thefootnote\relax\footnotetext{
$^*$ Work done during an internship at KwaiVGI, Kuaishou Technology. $^\dagger$ Corresponding authors
}

\vspace{-2em}
\begin{figure*}[!ht]
\centering
\includegraphics[width=\linewidth]{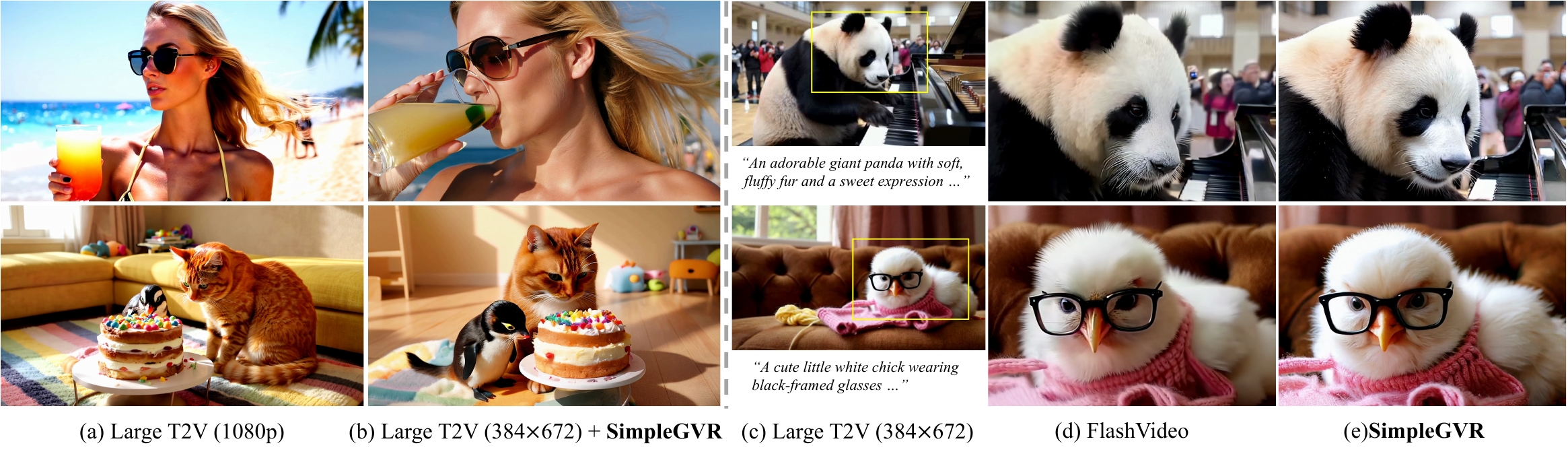}
\vspace{-8pt}
\caption{Built upon the low-resolution latent outputs (e.g., $384 \times 672$ resolution) from the first-stage Large T2V model, SimpleGVR generates high-quality results that even surpass the 1080p outputs of the Large T2V model. Compared to FlashVideo, which also adopts a cascaded architecture, SimpleGVR produces more realistic and finer details.
}
\label{fig:teaser}
\end{figure*}

\begin{abstract}
Latent diffusion models have emerged as a leading paradigm for efficient video generation. However, as user expectations shift toward higher-resolution outputs, relying solely on latent computation becomes inadequate. A promising approach involves decoupling the process into two stages: semantic content generation and detail synthesis. The former employs a computationally intensive base model at lower resolutions, while the latter leverages a lightweight cascaded video super-resolution (VSR) model to achieve high-resolution output. In this work, we focus on studying key design principles for latter cascaded VSR models, which are underexplored currently. First, we propose two degradation strategies to generate traininpairs that better mimic the output characteristics of the base model, ensuring alignment between the VSR model and its upstream generator. Second, we provide critical insights into VSR model behavior through systematic analysis of (1) timestep sampling strategies, (2) noise augmentation effects on low-resolution (LR) inputs. These findings directly inform our architectural and training innovations. Finally, we introduce interleaving temporal unit and sparse local attention to achieve efficient training and inference, drastically reducing computational overhead. Extensive experiments demonstrate the superiority of our framework over existing methods, with ablation studies confirming the efficacy of each design choice. Our work establishes a simple yet effective baseline for cascaded video super-resolution generation, offering practical insights to guide future advancements in efficient cascaded synthesis systems.
\end{abstract}

\section{Introduction}

Recent advancements~\cite{chen2023seine,fridman2024scenescape,yin2023nuwa,voleti2022mcvd,blattmann2023align,zhang2023show,bartal2024lumiere,jiang2023text2performer,polyak2410movie,ma2025step,kong2024hunyuanvideo,seawead2025seaweed,wan2025wan,yang2024cogvideox,zheng2024open} in diffusion-based text-to-video (T2V) generation have markedly enhanced the visual quality and coherence of synthesized videos. Leading models, such as Hunyuan~\cite{kong2024hunyuanvideo}, CogXVideo~\cite{yang2024cogvideox} and Wan~\cite{wan2025wan}, rely on DiT backbones with full self‑attention to fuse spatial, temporal, and textual cues, producing coherent clips with rich detail. However, their computational cost grows quadratically with spatial resolution: directly generating 1080p output in a single stage demands prohibitive computation and incurs long inference times. Given their multi‑resolution training paradigm, running these models at lower resolution (e.g., 512\textsuperscript{2}) strikes an optimal balance between visual fidelity and computational efficiency. 
We define 512p as a resolution whose total pixel area is nearly 512\textsuperscript{2}. Since 512p outputs capture motion and structure well, applying a lightweight cascaded VSR model to enhance them to 1080p offers a promising solution for efficient high-resolution video synthesis with low computational cost and latency.

Existing works, such as VEnhancer~\cite{he2024venhancer}, SeedVR~\cite{wang2025seedvr}, and FlashVideo~\cite{zhang2025flashvideo}, have demonstrated strong performance in video enhancement area. However, all these methods cannot operate directly on low-resolution latent representations. To apply these methods, users must first decode the low-resolution latent into a low-resolution video, apply upsampling, and then re-encode the video into a high-resolution latent before performing detail refinement. Consequently, this process introduces significant computational overhead. In this work, we propose SimpleGVR, the first method to support upsampling directly on the low-resolution latent representations produced by upstream T2V models, thereby eliminating redundant decoding and re-encoding steps and improving overall efficiency.

SimpleGVR adopts a simple yet effective architecture. It incorporates conditional information via concatenation and enables full-parameter finetuning. To improve SimpleGVR's performance, we focus on several key aspects. One aspect is the design of degradation pipeline, which is critical and has been extensively explored in real-world image and video restoration literature. In contrast, the degradation pipeline in AIGC remains underexplored. Unlike real-world degradations, which often stem from well-defined physical processes like image signal processing (ISP) pipelines, degradations in AIGC content lack clear physical analogs. Directly applying two-stage degradation strategies, like those in RealBasicVSR~\cite{realbasicvsr}, leads to severe artifacts and impaired depth perception when enhancing AIGC videos. To mimic the output characteristics of the computationally intensive base model (Large T2V model), we propose two degradation strategies: (1) Flow-based degradation, where optical flow guides motion-aware color blending and adaptive blurring, and (2) Model-guided degradation, where noise is added to low-resolution video frames and partially denoised using the base T2V model. These strategies generate training pairs that more faithfully reflect the characteristics of base T2V model output, narrowing the degradation gap compared to second-order degradation approach.

Another crucial aspect lies in the training configuration, which encompasses both timestep sampling strategy and noise augmentation to the low-resolution (LR) inputs. These components significantly affect SimpleGVR’s ability to enhance fine details and correct structural errors. Specifically, motivated by the observed differences in SimpleGVR’s capacity to reconstruct high-frequency details across different timestep, we propose a detail-aware sampler that outperforms the commonly used uniform sampler. Besides, the range of noise augmentation during training modulates the model's capacity for structural modification to the input LR, in addition to details generation. We find that using a middle-range noise level (e.g., $0.3\sim0.6$) strikes a good balance between enhancing high-frequency details and correcting structural errors in the input frames.

To enable SimpleGVR to handle more practical scenarios, such as processing 5-second videos (i.e., 77 frames), we further explore strategies for efficient training and inference. Due to GPU memory constraints, training directly on full 77-frame sequences is impractical. To address this, we adopt an efficient training strategy: starting with training on shorter 17-frame clips, then extending SimpleGVR to 77 frames through the interleaving temporal unit mechanism. Expanding SimpleGVR’s capability to handle 77 frames requires only 5K additional iterations. To support efficient inference, we further replace the full self-attention with the sparse local attention. This reduces computational cost of full self-attention by 80\% while maintaining comparable performance. Compared to swin attention~\cite{liu2021swin}, the sparse local attention mechanism achieves better detail reconstruction with less computation, striking a good balance between efficiency and quality.

In summary, our main contributions are as follows: \textbf{1)} We present SimpleGVR, the first lightweight model that performs video super-resolution directly on the latent representations of large T2V models. When applied to 512p outputs, SimpleGVR produces 1080p videos with higher quality than directly generated 1080p outputs from the same base Large T2V model. 
\textbf{2)} We design two degradation schemes, namely flow-based degradation and model-guided degradation synthesis, to simulate the degradation characteristics of the base model’s outputs. This ensures better alignment between the VSR model and its upstream generator. \textbf{3)} We analyze the timestep sampler and noise augmentation, and introduce a detail-aware sampler along with an appropriate noise augmentation range, which together enhance the model’s ability to recover fine details and correct structural errors. \textbf{4)} We introduce the interleaving temporal unit mechanism and incorporate the sparse local attention mechanism into SimpleGVR, enabling efficient training and inference while supporting long-sequence, high-resolution video generation. 

\section{Related Work}

\subsection{Cascade Diffusion Models.}

Cascade architectures have been widely explored in text-to-image and text-to-video generation~\cite{wang2025lavie,saharia2022image,zhang2025flashvideo}, where multi-stage designs are employed to address the challenge of generating high-resolution outputs. Typically, a low-resolution sample is first generated, followed by a specialized model to progressively refine details. Among these, FlashVideo~\cite{zhang2025flashvideo} is most related to our work. It begins second-stage generation from low-quality video inputs rather than pure guassian noise, enabling efficient high-resolution synthesis with only 4 function evaluations. However, SimpleGVR differs in two key aspects. First, we concatenate the low-resolution latent as a condition, allowing the model not only to leverage coarse content but also to correct structural errors within it. This design leads to better performance under the same 50-step inference setting. Second, we introduce two degradation strategies that explicitly simulate the characteristic degradations from the first-stage T2V generator.

\subsection{Degradation Models in Restoration.}

Degradation modeling is important for effective image and video restoration. Traditional models~\cite{dong2014learning,dong2016accelerating,gu2019ikc} often rely on simple assumptions like bicubic downsampling or gaussian blur, which fail to capture complex real-world degradations. Prior works such as BSRGAN~\cite{zhang2021designing}, Real-ESRGAN~\cite{wang2021real}, and video-oriented methods like RealBasicVSR~\cite{realbasicvsr} simulate more complicated degradations, including blur, noise, and compression, improving robustness on real-world low quality images and videos. However, these models are designed for real-world scenarios and do not account for the unique distortions in AIGC-generated videos, such as motion blur and color blending. These AIGC-specific artifacts require specialized degradation modeling. To this end, we propose a heuristic degradation strategy consisting of two complementary components: a flow-based degradation scheme and a model-guided degradation scheme via SDEdit. Together, they enable the generation of training pairs that more faithfully mimic the output characteristics of the first-stage T2V generator.

\subsection{Video Restoration.}

Early video restoration (VR) methods~\cite{chan2021basicvsr,chan2022basicvsr,chen2024learning,li2023baseliner,liang2022recurrent,wang2019edvr,youk2024fma} rely on synthetic data, limiting real-world performance. Later works~\cite{realbasicvsr,xie2023mitigating,zhang2024realviformer} shift toward real scenarios but still struggle with texture realism. Diffusion-based approaches~\cite{he2024venhancer,wang2025lavie,li2025diffvsr,wang2025seedvr,zhang2025flashvideo} leverage generative priors to achieve more realistic and coherent video restoration. However, all these methods require decoded RGB frames and cannot operate directly on latent representations, making them less suitable for T2V pipelines. In contrast, our SimpleGVR performs upsampling and refinement directly in the latent space of the upstream generator, enabling seamless integration with generative video models. Unlike SeedVR, which uses a fixed window attention mechanism, SimpleGVR adopts a more advanced local attention strategy that achieves better performance under the lower computational budget. 

\section{Preliminary} 
SimpleGVR is conducted over an internal pre-trained text-to-video foundation model, which is composed of a 3D Variational Autoencoder (VAE)~\cite{kingma2013vae}, a T5-based text encoder~\cite{raffel2020t5}, and a transformer-based latent diffusion module~\cite{chen2023pixart,peebles2023dit}. The generative transformer operates over latent representations and is structured with a repeated stack of components: 2D spatial self-attention, 3D spatiotemporal attention, text-guided cross-attention, and feed-forward layers. The input text prompt is encoded by the T5 model into a conditioning vector $c_{text}$, which guides the generation model. We follow the Rectified Flow framework~\cite{esser2024rectified} to define a linear path between the clean latent $z_{0}$ and its noisy counterpart $z_{t}$ at timestep $t$:

\begin{equation}
z_t = (1 - t) z_0 + t \epsilon, \quad \epsilon \sim \mathcal{N}(0, I)
\end{equation}

The denoising process is formulated as ordinary differential equation (ODE) that maps $z_{t}$ back to $z_{0}$:

\begin{equation}
d z_t = v_\Theta(z_t, t, c_{\text{text}}){d t},
\end{equation}
where the velocity field $v$ is modeled by a neural network with parameters $v_\Theta$. During training, Conditional Flow Matching (CFM)~\cite{lipman2022flowmatching} is used to regress the velocity via the following objective:

\begin{equation}
\mathcal{L}_{\text{CFM}} = \mathbb{E}_{t, \epsilon \sim \mathcal{N}(0, I), z_0} 
\left[ \left\| (z_1 - z_0) - v_\Theta(z_t, t, c_{\text{text}}) \right\|_2^2 \right].
\end{equation}

\section{Methodology}

Our cascaded video generation framework operates within a latent space defined by a pre-trained VAE. The framework comprises two core components:
(i) A computationally intensive base Text-to-Video (T2V) model, which employs a Diffusion Transformer (DiT~\cite{peebles2023scalable}) architecture to generate low-resolution video latent representations.
(ii) A cascaded latent video super-resolution model, termed SimpleGVR, which adopts a lightweight architecture to efficiently enhance the base model’s output into high-resolution video latent representations. The overall framework structure is illustrated in Fig.~\ref{fig:pipeline}(b). As the primary focus of this paper, our method addresses the task of the latter component.

In the following sections, we first formalize the problem addressed by SimpleGVR (Sec.~\ref{sec:pipeline}). We then investigate this critical yet underexplored challenge through three key perspectives:
Degradation simulation for synthesizing training pairs (Sec.~\ref{sec:degradation});
Training configurations to promote faithful detail generation (Sec.~\ref{sec:analysis});
Efficient designs to address high-resolution video computation demands (Sec.~\ref{sec:efficient}). Note that the 512p is defined as a resolution whose total pixel area is approximately 512\textsuperscript{2}.

\subsection{Formulation of SimpleGVR}
\label{sec:pipeline}
\begin{figure*}[t]
\centering
\small 
\begin{minipage}[t]{0.95\linewidth}
\centering
\includegraphics[width=1.0\columnwidth]{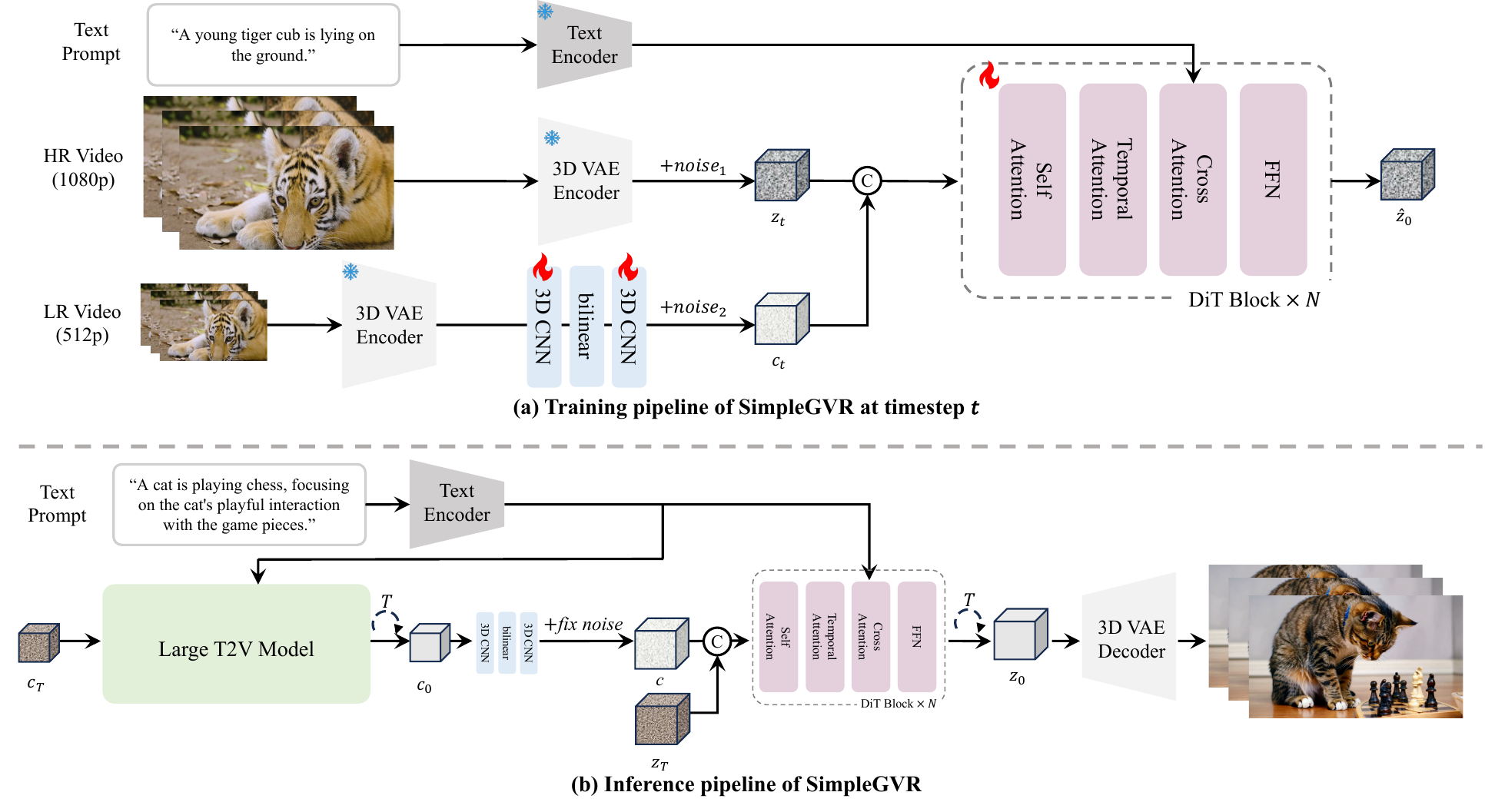}
\end{minipage}
\centering
\caption{The training and inference pipeline of SimpleGVR. Unlike previous methods, SimpleGVR performs upsampling directly at the latent level. This design allows the latent representation $c_{0}$ produced by the upstream Large T2V model to be directly processed by SimpleGVR during inference, eliminating redundant decoding and re-encoding steps.} 
\label{fig:pipeline}
\end{figure*}

Fig.~\ref{fig:pipeline} (a) and (b) illustrate the training and inference pipelines of SimpleGVR, respectively. To optimize SimpleGVR, we adopt a diffusion model. During training, the 512p low-resolution (LR) video and the corresponding 1080p high-resolution (HR) video are first encoded into latent representations via a 3D VAE~\cite{kingma2013auto}, yielding LR and HR latents. To align the spatial dimensions of the LR latent with the HR latent, we apply a 3D CNN followed by a bilinear upsampling operator, and then another 3D CNN layer. Two independent random noises are then injected into both latents with different magnitudes, yielding noisy representations $c_{t}$ and $z_{t}$. Here, \( z_t \) refers to the noisy latent in the diffusion process, while \( c_t \) denotes the noisy LR latent that serves as the conditioning input. To distinguish the roles of these two noise injections, we refer to the perturbation applied to the LR latent as noise augmentation~\cite{he2024venhancer,wang2025seedvr}, which plays a crucial role in enhancing the model’s capacity for detail generation and correcting structural errors. This effect is further analyzed in later sections. The noisy LR and HR latents are then fused via channel-wise concatenation and fed into a series of DiT blocks. All parameters, including those of the 3D CNN and DiT blocks, are optimized jointly in an end-to-end fashion during training.

Once trained, SimpleGVR can be directly applied to the low-resolution latents generated by the large T2V model. Specifically, given a random gaussian noise $c_{T}$, the large T2V model performs multiple denoising steps to produce a clean low-resolution latent $c_{0}$. This latent is then upsampled and perturbed with a fixed level of random noise to yield $c$. Concurrently, a high-resolution gaussian noise sample $z_{T}$ is randomly initialized. The noisy high-resolution latent $z_{T}$ and the conditioned latent $c$ are concatenated along the channel dimension and fed into the DiT blocks of SimpleGVR. Notably, the conditioning latent $c$ remains fixed throughout the denoising process. After the iterative refinement, the final clean high-resolution latent $z_{0}$ is decoded to obtain a high-fidelity 1080p video.

\subsection{Degradation Modeling}
\label{sec:degradation}

\begin{figure*}[t]
\centering
\small 
\begin{minipage}[t]{0.9\linewidth}
\centering
\includegraphics[width=1.0\columnwidth]{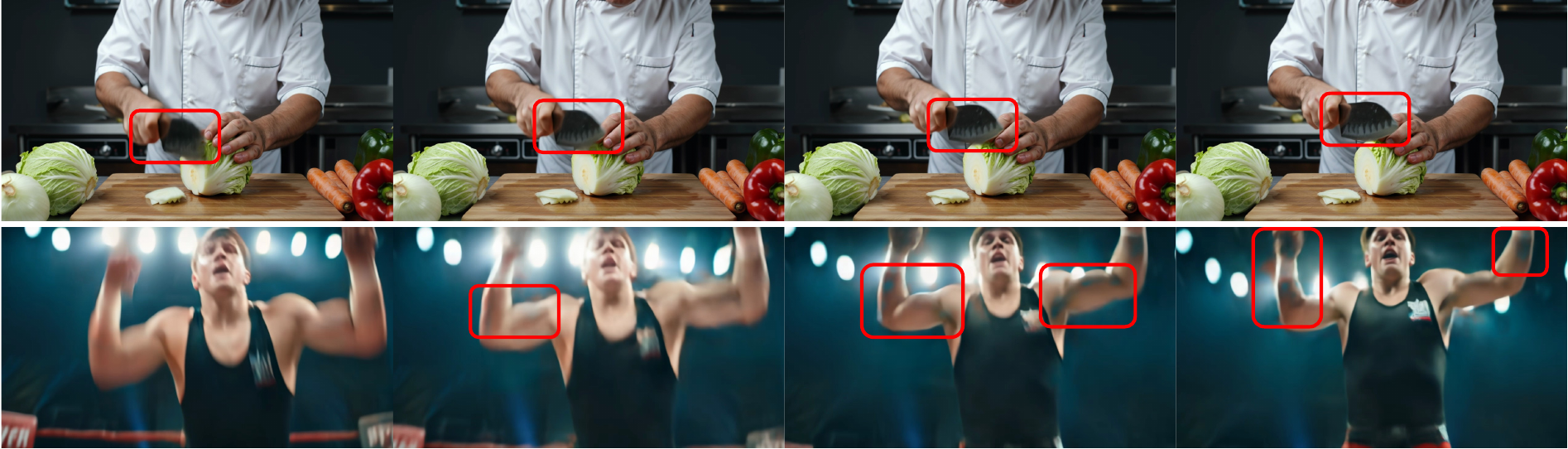}
\end{minipage}
\centering
\caption{Visual artifacts in decoded videos from the Large T2V model. Dynamic regions exhibit noticeable local motion blur and color blending distortions.} 
\label{fig:large_motion}
\end{figure*}

\begin{figure}[t]
\centering
\small 
\begin{minipage}[t]{1.0\linewidth}
\centering
\includegraphics[width=1.0\columnwidth]{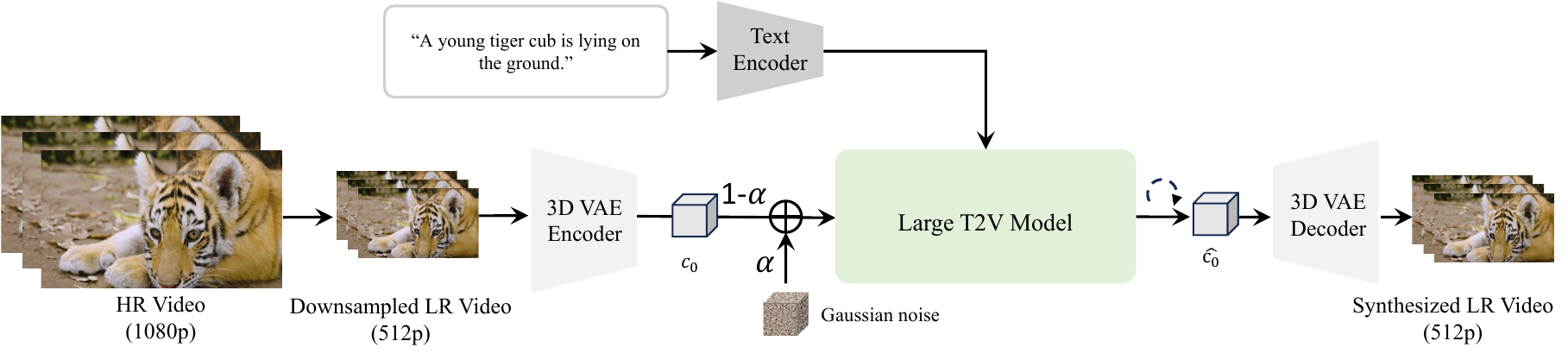}
\end{minipage}
\centering
\caption{Model-guided degradation synthesis pipeline. The parameter $\alpha$ controls the strength of the added Gaussian noise, which also affects the structural alignment between $\hat{c}_{0}$ and $c_{0}$.} 
\label{fig:sdedit}
\end{figure}

\subsubsection{Flow-based degradation} 

Upon inspecting the first-stage T2V outputs (see Fig.~\ref{fig:large_motion}), we observe that unlike real-world low-quality videos, these video sequences do not exhibit severe degradations such as severe blur, noise, or compression. Instead, they present two distinctive characteristics: 1) localized motion‑dependent blur and (2) frame‑to‑frame color blending distortion (current‑frame colors smeared by previous‑frame hues). These two phenomena are closely entangled and cannot be effectively replicated using traditional degradation model~\cite{realbasicvsr}. 
To synthesize color blending distortion,  we first estimate the motion field between adjacent frames using Dense Inverse Search (DIS)~\cite{dis} optical flow algorithm, mapping each point $(x,y)$ in the current frame to its position $(x',y')$ in the previous frame. Based on the optical flow results, we generate motion masks to identify regions with significant movement and create elliptical patterns within these regions.  For each elliptical region, we apply gaussian sampling to extract colors from the previous frame, blend them in RGB space, and generate the colors. We then blend these colors into the current frame using a distance-based weight map, where pixels closer to the ellipse center receive stronger effects.

After applying color blending, we synthesize motion blur effects on the processed frames. Based on the motion field computed earlier, for each block in the frame, we generate adaptive blur kernels whose parameters are determined by the local motion characteristics. The kernel size and shape vary according to the motion magnitude, with larger motion leading to longer blur kernels aligned with the motion direction. We apply these kernels through a weighted convolution operation, where the blur intensity is proportional to the local motion magnitude. This block-wise processing ensures that static regions maintain their sharpness while areas with significant movement exhibit realistic motion blur.

\subsubsection{Model-guided degradation}

SimpleGVR is fundamentally designed to learn a mapping from the output domain of large T2V models to high-quality video data. By constructing paired training samples where the low-resolution inputs are directly sourced from the T2V model outputs, SimpleGVR can be better aligned with the distribution and artifacts specific to the T2V model. Inspired by SDEdit~\cite{meng2021sdedit}, which injects noise into the input and leverages a diffusion model as the structural prior to produce outputs that are both visually realistic and structurally faithful, we adopt a similar approach in our framework. As shown in Fig.~\ref{fig:sdedit}, we begin by downsampling a high-quality 1080p video to 512p and encoding it via a 3D VAE to obtain the latent $c_{0}$. This latent is blended with a gaussian noise under a predefined ratio $\alpha$, and the noisy latent is then partially denoised using the large T2V model to generate $\hat{c}_{0}$.
A higher $\alpha$ pushes $\hat{c}_{0}$ closer to the T2V distribution but weakens its structural alignment with the original video. To balance realism and fidelity, we set $\alpha \in [0.3, 0.4]$, ensuring that $\hat{c}_0$ retains the overall layout of the source video while approximating the output domain of the Large T2V model. The synthesized LR video produced by the 3D VAE encoder can be used as the LR input during training (see Fig.~\ref{fig:pipeline}(a)). In practice, to reduce the overhead of decoding and re-encoding, we store only the latent representation $\hat{c}_0$, which is directly used as input to the 3D CNN.

\subsection{Training Configuration}
\label{sec:analysis}

To further enhance SimpleGVR’s ability to recover fine details and correct structural errors in the input frames, we analyze two key training configurations: the timestep sampling scheduling and the noise augmentation applied to the low-resolution (LR) branch.

\begin{figure}[htbp]
\centering
\begin{minipage}{0.48\textwidth}
    \centering
    \includegraphics[width=\textwidth]{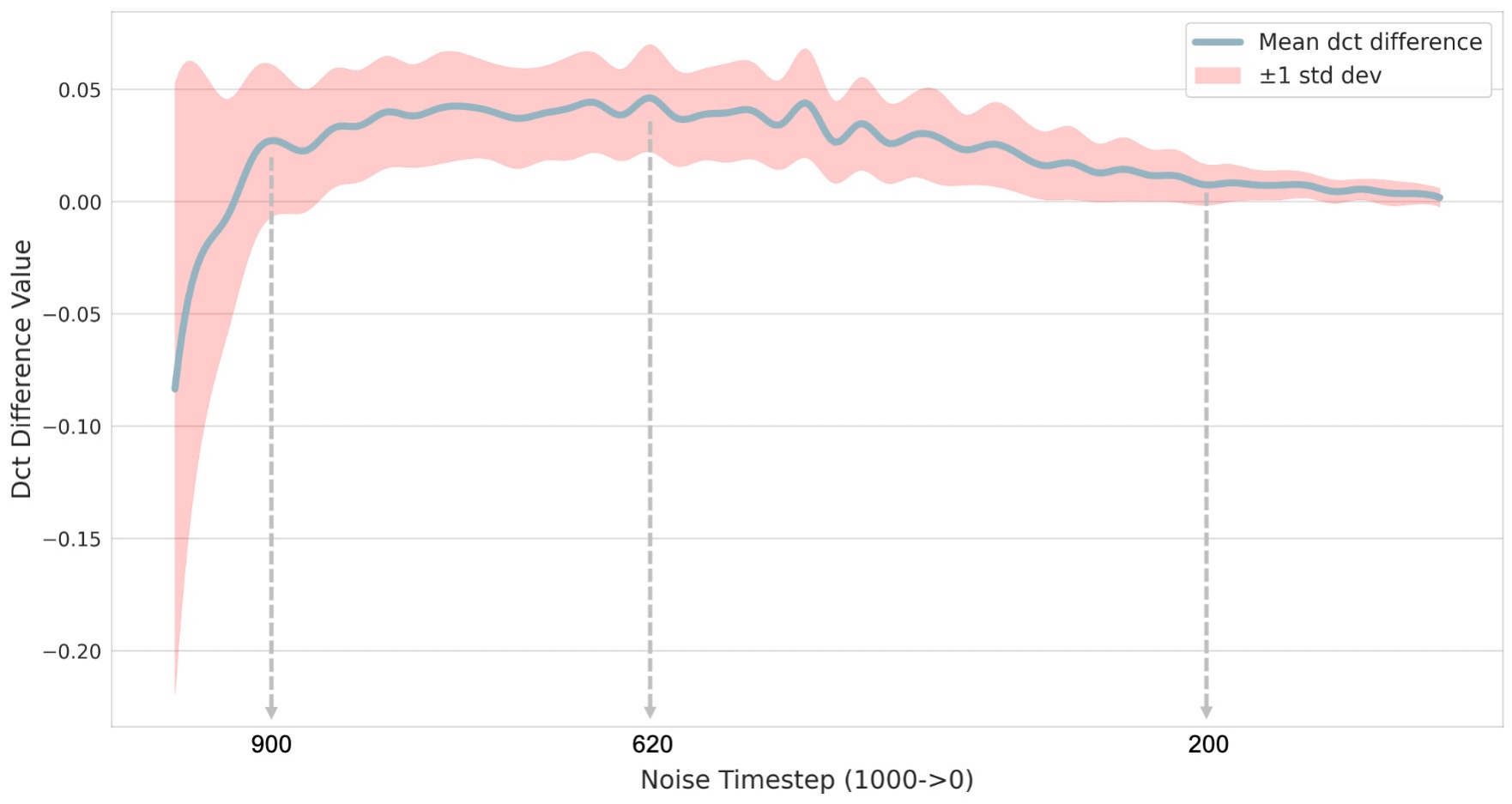}
    \caption{High-frequency variation curve over timesteps during inference.} 
    \label{fig:sampler}
\end{minipage}
\hspace{0.02\textwidth} %
\begin{minipage}{0.48\textwidth}
    \centering
    \resizebox{\textwidth}{!}{%
    \begin{tabular}{c|c|ccc}
    \hline
    {\color[HTML]{262626} }                          & {\color[HTML]{262626} }                        & \multicolumn{3}{c}{{\color[HTML]{262626} DOVER}}                                                     \\ \cline{3-5} 
    \multirow{-2}{*}{{\color[HTML]{262626} Sampler}} & \multirow{-2}{*}{{\color[HTML]{262626} MUSIQ}} & {\color[HTML]{262626} Technical} & {\color[HTML]{262626} Aesthetic} & {\color[HTML]{262626} Overall} \\ \hline
    Uniform                                          & {\color[HTML]{262626} 62.04}                   & {\color[HTML]{262626} 18.58}     & {\color[HTML]{262626} 98.78}     & {\color[HTML]{262626} 68.94}   \\ \hline
    Detail-aware                                     & {\color[HTML]{262626} 62.19}                   & {\color[HTML]{262626} 18.92}     & {\color[HTML]{262626} 98.83}     & {\color[HTML]{262626} 69.64}   \\ \hline
    \end{tabular}
    }
    \caption{Quantitative comparison between the uniform sampler and the detail-aware sampler on the AIGC100 dataset, demonstrating that the detail-aware sampler outperforms the uniform sampler in most metrics. These experiments are conducted on 17-frame inputs for 20K iterations.}
    \label{tab:sampler}
\end{minipage}%
\end{figure}

\paragraph{Timestep Sampling Scheduling.}
The uniform sampler, where each timestep is sampled with equal probability, is commonly used for training diffusion-based models. However, since SimpleGVR focuses on detail synthesis, understanding which timesteps contribute most to enhancing visual details during denoising is crucial. To this end, we analyze high-frequency detail changes in the predicted $\hat{z}_t^0$ at each denoising step. Specifically, we sample 200 low-resolution 512p test videos and perform 50-steps inference using the SimpleGVR model trained with a uniform sampler. At each denoising timestep $t$, we obtain the latent $z_t$ and directly predict its corresponding clean signal $\hat{z}_t^0$. To quantify the high-frequency content of $\hat{z}_t^0$, we apply discrete cosine transform (DCT) and extract its high-frequency coefficients $\mathcal{H}(\hat{z}_t^0)$. We then compute the pairwise differences of these high-frequency components across timesteps to derive the detail variation curve shown in Fig.~\ref{fig:sampler}.
The figure shows that significant detail gains primarily occur in the high-noise and mid-noise regions, while the low-noise region contributes minimally. Based on this observation, we normalize the weights to create a probability distribution, which we then adopt as the detail-aware sampler. Replacing the uniform sampler with this detail-aware sampler during training leads to improved performance, as evidenced by the results in Table~\ref{tab:sampler}.

\begin{figure*}[t]
\centering
\small 
\begin{minipage}[t]{0.95\linewidth}
\centering
\includegraphics[width=1.0\columnwidth]{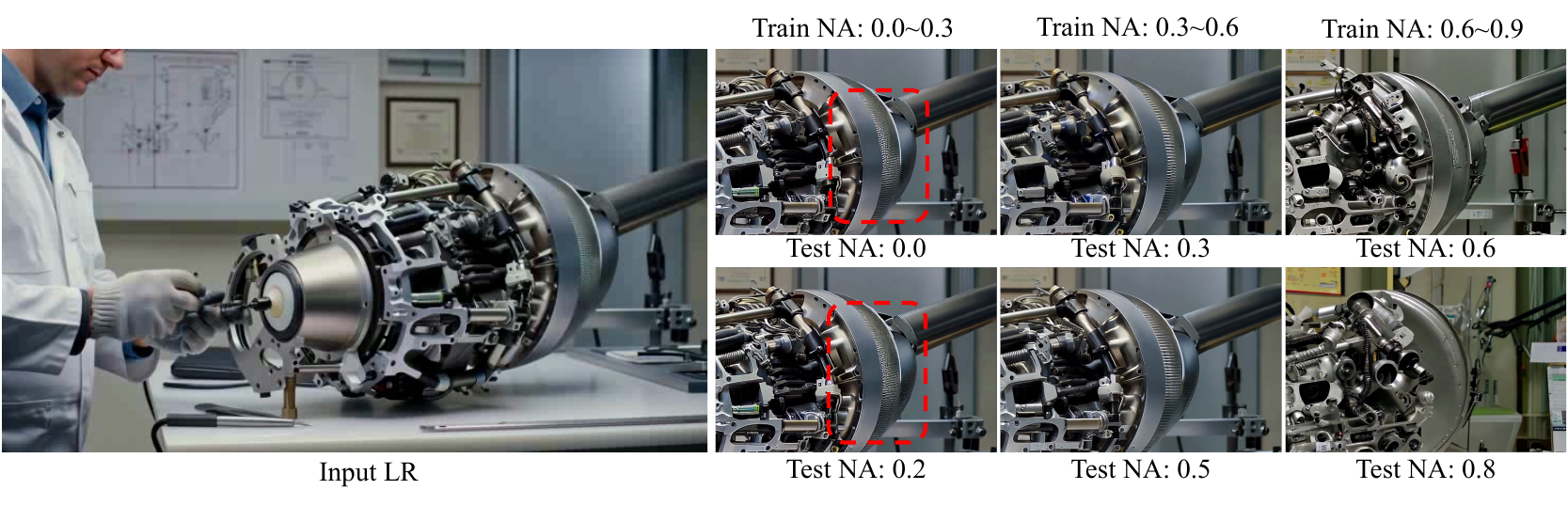}
\end{minipage}
\centering
\caption{Visual results of SimpleGVR trained with different noise augmentation (NA) ranges. When the NA during training is small, SimpleGVR struggles to remove artifacts from the low-resolutn video. When the NA is large, SimpleGVR has difficulty preserving the original structure of the low-resolution video. Only when the NA is in a moderate range does SimpleGVR strike a good balance between correcting structural errors and enhancing high-frequency details.} 
\label{fig:noise_aug}
\end{figure*}

\paragraph{Noise augmentation Effect.}
As noted in prior work~\cite{he2024venhancer,wang2025seedvr,zhang2025flashvideo}, introducing noise in the LR branch during training not only enhances the generative capacity of video enhancement models but also facilitates controllable behavior at inference time. To identify a suitable noise range for SimpleGVR, we divide the noise scale into three intervals: small ($0.0\sim0.3$), middle ($0.3\sim0.6$), and large ($0.6\sim0.9$). By training and testing the model separately within each interval, we observe that different noise levels lead to distinct behaviors. When trained with noise in the small interval, the model exhibits limited ability in refining fine details. In particular, when the input video frames contain structural errors, increasing the noise level in the LR branch tends to introduce more artifacts rather than meaningful corrections, as illustrated by the red box in Fig.~\ref{fig:noise_aug}. In contrast, training with large noise results in significant deviations in structural content, causing the output to diverge from the input. Notably, only the middle interval strikes a good balance: the model is capable of enhancing high-frequency details while still being able to correct structural errors in the input frames.

\subsection{Efficient Pseudo-Global Computation}
\label{sec:efficient}

\paragraph{Interleaving Temporal Unit.}

\begin{wrapfigure}{r}{0.5\textwidth} %
\centering
\vspace{-40pt}
\includegraphics[width=0.48\textwidth]{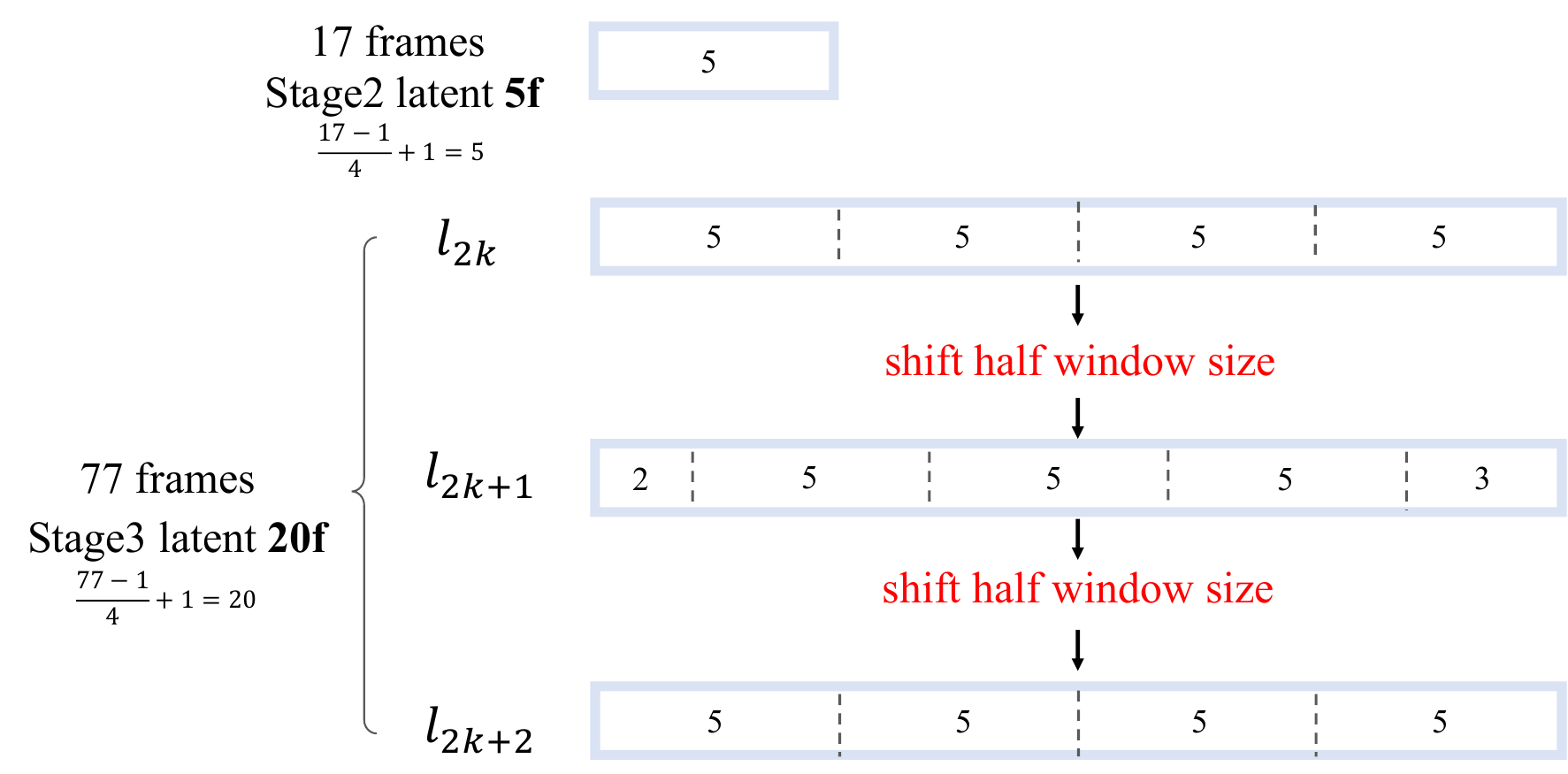} %
\caption{Visualization of the interleaving temporal unit mechanism.}
\label{fig:temporal_unit}
\vspace{-4pt}
\end{wrapfigure}

Fig.~\ref{fig:temporal_unit} illustrates the design of the interleaving temporal unit mechanism for a 77-frame input. When the input frame count is 17 or 77, the latent vectors obtained via the VAE are 5 and 20 along the time dimension, respectively. Due to GPU memory limitations, we are unable to process the entire 77-frame data at once. Therefore, we first train a SimpleGVR model capable of handling 17 frames, and then extend it to 77 frames. During this extension, we load all 77 frames at once, then the corresponding latent is sliced along the time dimension into units of 5, with attention computations performed within each unit. To facilitate information exchange between units, inspired by Swin Attention~\cite{liu2021swin}, we apply a shift operation on the entire latent at the layer $l_{2k+1}$, and then perform another shift at the layer $l_{2k+2}$. Through this interleaved temporal unit design, we achieve highly efficient training, and the final model can effectively handle 77-frame video sequences.

\paragraph{Sparse Local Attention.}
To reduce computational overhead during inference, we further explore replacing the full attention in SimpleGVR with swin attention. However, despite the use of the shifted window mechanism, this modification leads to noticeable reduction in fine details.
We hypothesize that this is due to its limited receptive field and lack of dynamic cross-window interactions. 
Inspired by MoBA~\cite{lu2025moba} and swin attention's window partitioning strategy, we divide tokens into 2D window units of fixed size. Each unit performs self-attention within itself and dynamically attends to the top‑k most relevant window units across the entire video, including future frames, based on computed relevance scores. This mechanism enables effective long-range modeling and detail preservation, even with small window sizes and small top‑k values (e.g., 1).

\section{Experiments}

\subsection{Implementation Details}

\subsubsection{Training Dataset} 
We collect approximately 840K high-quality
video clips from the Internet to construct our training dataset. Given the highly variable quality of online videos, we design an automated filtering pipeline to retain only visually high-quality content. Specifically, we first discard videos that are overly bright or dark. Then, for each video, we uniformly sample 10 frames and compute two
metrics: the average MUSIQ score~\cite{ke2021musiq} and the Laplacian variance, which reflects the level of spatial detail or sharpness.
Videos with an average MUSIQ score below 40 or a Laplacian variance below 30 are discarded.

\subsubsection{Testing Dataset}
To ensure diversity in our test samples, we first decode the latent representations from the T2V base model into pixel space. Based on this output, we collect a set of 100 video clips across diverse scenarios, including human, animal, object, motion, and defocus scenes. We refer to this curated set as AIGC100, with each clip containing 77 frames.

\subsubsection{Metrics}
Since AIGC100 is generated by the first-stage Large T2V model, ground-truth references are not available for full-reference evaluation. Consequently, we adopt a suite of no-reference metrics to assess both frame-level and video-level quality. Specifically, we use MUSIQ~\cite{ke2021musiq} for single-frame perceptual quality, DOVER~\cite{wu2023exploring} for overall video quality, and a set of metrics from VBench~\cite{huang2024vbench}, which evaluate diverse aspects of AIGC videos, including background consistency, subject consistency, aesthetic quality, imaging quality, and motion smoothness.

\subsubsection{Training Details}

SimpleGVR is trained on 16 NVIDIA H800 GPUs (80GB each) with a total batch size of 32. We use the AdamW optimizer~\cite{adamw} with a learning rate of $5\times10^{-5}$, and randomly replace the text prompt with a null prompt in 10\% of cases to enhance robustness. The training pipeline is divided into three stages. In the first stage, initialized from a pretrained 1B T2V model, SimpleGVR is trained for 20K iterations on 17-frame inputs using training pairs constructed via a one-order degradation process. In the second stage, we fine-tune the model for an additional 5K iterations on a dataset (30K) generated using the proposed degradation strategies. In the third stage, based on the dataset synthesized in the previous two stages, we continue fine-tuning SimpleGVR and extend the temporal range to 77 frames by using the temporal window attention training strategy. Note that we also implement the sparse local attention mechanism in this stage.
During the whole training pipeline, the LR branch is injected with noise sampled from the range $[0.3, 0.6]$, corresponding to diffusion timesteps randomly selected from $[300, 600]$.

\subsection{Comparison with SOTA methods}
\begin{figure*}[htbp]
\centering
\small 
\begin{minipage}[t]{1.0\linewidth}
\centering
\includegraphics[width=1.0\columnwidth]{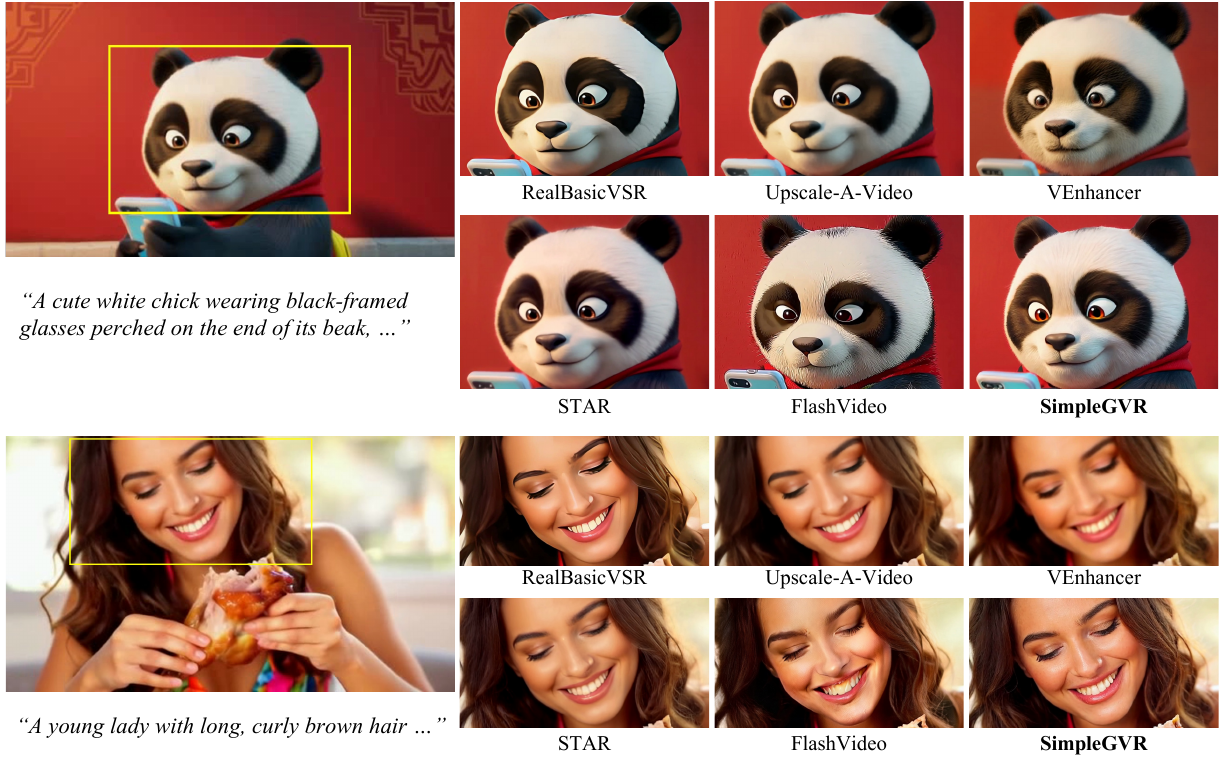}
\end{minipage}
\centering
\caption{Qualitative comparisons on AIGC100 dataset. Our SimpleGVR is capable of generating more realistic details than our methods.} 
\label{fig:visual_comparison}
\end{figure*}

We compare SimpleGVR with existing state-of-the-art methods, RealBasicVSR~\cite{realbasicvsr}, Upscale-A-Video~\cite{uav}, VEnhancer~\cite{he2024venhancer}, STAR~\cite{xie2025star} and FlashVideo~\cite{zhang2025flashvideo}. For fair comparison, we set the inference steps of FlashVideo to 50. As shown in Table~\ref{tab:comparison}, SimpleGVR achieves the best performance on both MUSIQ and DOVER. Moreover, regarding the comprehensive metrics proposed in VBench, SimpleGVR also achieves the highest average score. Qualitative comparisons are presented in Fig.~\ref{fig:visual_comparison}. Compared to other methods, SimpleGVR adds realistic details while maintaining the original style and semantics. For example, while VEnhancer introduces unnatural textures around the panda’s eyes and alters the style, SimpleGVR maintains more natural details. Similarly, for human faces, SimpleGVR produces finer and more realistic details. In contrast, other methods either struggle to generate sufficient detail or create noticeable artifacts, resulting in outputs less realistic than those produced by SimpleGVR.

\begin{table}[H]
\caption{Quantitative comparison on AIGC100 dataset. \textbf{Bold} and \underline{underline} indicate the best and second best performance.}
\centering
\resizebox{1.0\textwidth}{!}{
\begin{tabular}{c|c|ccc|cccccc}
\hline
{\color[HTML]{262626} }                         & {\color[HTML]{262626} }                        & \multicolumn{3}{c|}{{\color[HTML]{262626} DOVER}}                                                                     & \multicolumn{6}{c}{{\color[HTML]{262626} Vbench Metrics}}                                                                                                   \\ \cline{3-11} 
\multirow{-2}{*}{{\color[HTML]{262626} Method}} & \multirow{-2}{*}{{\color[HTML]{262626} MUSIQ}} & {\color[HTML]{262626} Technical}      & {\color[HTML]{262626} Aesthetic}      & {\color[HTML]{262626} Overall}        & {\color[HTML]{262626} \begin{tabular}[c]{@{}c@{}}Background\\ Consistency\end{tabular}} & {\color[HTML]{262626} \begin{tabular}[c]{@{}c@{}}Subject \\ Consistency\end{tabular}} & {\color[HTML]{262626} \begin{tabular}[c]{@{}c@{}}Aesthetic\\ Quality\end{tabular}} & {\color[HTML]{262626} \begin{tabular}[c]{@{}c@{}}Imaging\\ Quality\end{tabular}} & {\color[HTML]{262626} \begin{tabular}[c]{@{}c@{}}Motion\\ Smoothness\end{tabular}} & \begin{tabular}[c]{@{}c@{}}Average \\ Score\end{tabular} \\ \hline
{\color[HTML]{262626} RealBasicVSR}             & {\color[HTML]{262626} {\underline{57.55}}}             & {\color[HTML]{262626} 12.27}          & {\color[HTML]{262626} {\underline{98.66}}}    & {\color[HTML]{262626} 61.84}          & {\color[HTML]{262626} 93.73}                                                            & {\color[HTML]{262626} 93.98}                                                          & {\color[HTML]{262626} 61.63}                                                       & {\color[HTML]{262626} 72.76}                                                     & {\color[HTML]{262626} 98.70}                                                       & {\underline{84.16}}                                              \\
{\color[HTML]{262626} VEnhancer}                & {\color[HTML]{262626} 40.03}                   & {\color[HTML]{262626} 15.38}          & {\color[HTML]{262626} 98.32}          & {\color[HTML]{262626} 62.54}          & {\color[HTML]{262626} 94.59}                                                            & {\color[HTML]{262626} 94.44}                                                          & {\color[HTML]{262626} 59.98}                                                       & {\color[HTML]{262626} 64.22}                                                     & {\color[HTML]{262626} 99.16}                                                       & 82.48                                                    \\
{\color[HTML]{262626} Upscale-A-Video}          & {\color[HTML]{262626} 36.35}                   & {\color[HTML]{262626} 12.43}          & {\color[HTML]{262626} 98.29}          & {\color[HTML]{262626} 59.04}          & {\color[HTML]{262626} 95.96}                                                            & {\color[HTML]{262626} 94.41}                                                          & {\color[HTML]{262626} 61.26}                                                       & {\color[HTML]{262626} 63.85}                                                     & {\color[HTML]{262626} 98.99}                                                       & 82.89                                                    \\
{\color[HTML]{262626} STAR}                     & {\color[HTML]{262626} 46.73}                   & {\color[HTML]{262626} {\underline{18.17}}}    & {\color[HTML]{262626} {\underline{98.66}}}    & {\color[HTML]{262626} {\underline{67.76}}}    & {\color[HTML]{262626} 96.17}                                                            & {\color[HTML]{262626} 94.43}                                                          & {\color[HTML]{262626} 62.24}                                                       & {\color[HTML]{262626} 67.24}                                                     & {\color[HTML]{262626} 99.01}                                                       & 83.82                                                    \\
{\color[HTML]{262626} Flashvideo}               & {\color[HTML]{262626} 53.65}                   & {\color[HTML]{262626} 15.97}          & {\color[HTML]{262626} 98.61}          & {\color[HTML]{262626} 65.38}          & {\color[HTML]{262626} 95.49}                                                            & {\color[HTML]{262626} 94.75}                                                          & {\color[HTML]{262626} 60.76}                                                       & {\color[HTML]{262626} 69.11}                                                     & {\color[HTML]{262626} 98.45}                                                       & 83.71                                                    \\ \hline
\textbf{Ours}                                   & {\color[HTML]{262626} \textbf{62.35}}          & {\color[HTML]{262626} \textbf{20.44}} & {\color[HTML]{262626} \textbf{98.88}} & {\color[HTML]{262626} \textbf{71.34}} & {\color[HTML]{262626} 95.35}                                                            & {\color[HTML]{262626} 94.32}                                                          & {\color[HTML]{262626} 62.84}                                                       & {\color[HTML]{262626} 71.91}                                                     & {\color[HTML]{262626} 98.74}                                                       & \textbf{84.63}                                           \\ \hline
\end{tabular}
}
\label{tab:comparison}
\end{table}

\subsection{Ablation Study}
\subsubsection{Effectiveness of Degradation Strategies}

\begin{figure}[H]
\centering
\small 
\begin{minipage}[t]{0.70\linewidth}
\centering
\includegraphics[width=1.0\columnwidth]{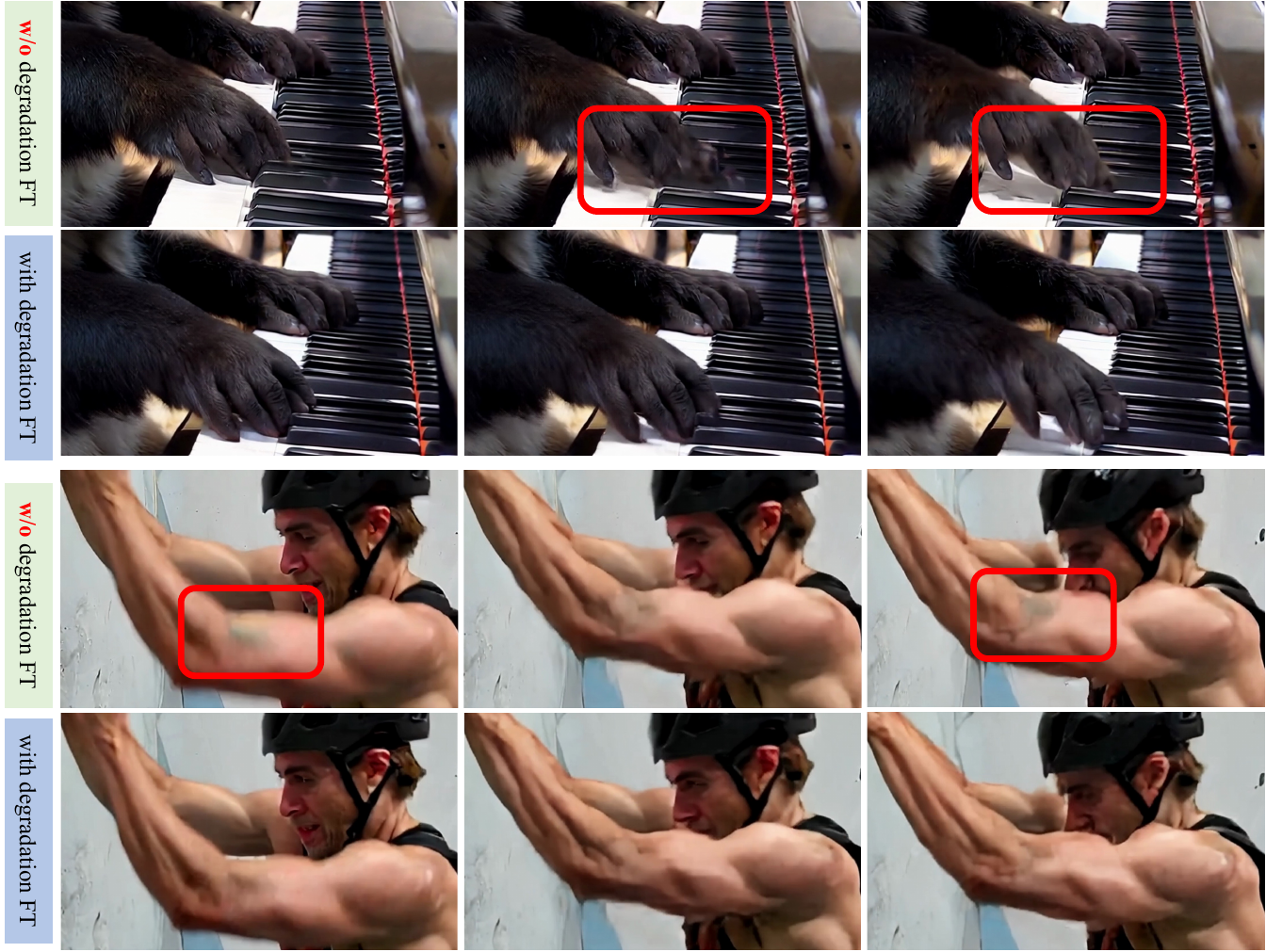}
\end{minipage}
\centering
\caption{Visualization of three consecutive frames generated by SimpleGVR. "w/o degradation FT" indicates SimpleGVR is trained only with one-order degradation, without fine-tuning using the proposed degradation strategies.} 
\label{fig:degradation}
\end{figure}

In Fig.~\ref{fig:degradation}, we demonstrate the effectiveness of the proposed degradation strategies. As shown in the first and third rows, the SimpleGVR trained solely with one-order degradation exhibits noticeable temporal inconsistencies in motion areas, such as the panda’s paw, and suffers from color blending artifacts, particularly evident in the human arm. After fine-tuning with training pairs generated by the two proposed degradation schemes, SimpleGVR effectively mitigates abrupt changes in motion regions across consecutive frames and significantly reduces color blending distortions.

\subsubsection{Effectiveness of Sparse Local Attention} To verify the effectiveness of sparse local attention, we train SimpleGVR with different attention operations on 17-frame inputs for 20K iterations. As shown in Table~\ref{tab:moba}, compared to Swin Attention, we observe that sparse local attention achieves better performance with lower computational cost. In our setting, the block size is set to $12\times9$, and each block attends to only one additional neighboring block. Under this configuration, sparse local attention reduces computational cost by approximately 80\% compared to full attention, while maintaining nearly the same performance. These results show the promising potential of sparse local attention mechanisms for video super-resolution tasks.

\begin{table}[ht]
\centering
\caption{Computational cost and performance comparison.}
\resizebox{0.70\textwidth}{!}{%
\begin{tabular}{c|c|c|ccc}
\hline
{\color[HTML]{262626} }                                     & {\color[HTML]{262626} }                            & {\color[HTML]{262626} }                        & \multicolumn{3}{c}{{\color[HTML]{262626} DOVER}}                                                     \\ \cline{4-6} 
\multirow{-2}{*}{{\color[HTML]{262626} Attention Operator}} & \multirow{-2}{*}{{\color[HTML]{262626} FLOPS (G)}} & \multirow{-2}{*}{{\color[HTML]{262626} MUSIQ}} & {\color[HTML]{262626} Technical} & {\color[HTML]{262626} Aesthetic} & {\color[HTML]{262626} Overall} \\ \hline
{\color[HTML]{262626} Full Attention}                       & {\color[HTML]{262626} 1612.5}                      & {\color[HTML]{262626} 62.19}                        & {\color[HTML]{262626} 18.92}     & {\color[HTML]{262626} 98.83}     & {\color[HTML]{262626} 69.64}   \\
{\color[HTML]{262626} Swin Attention}                       & {\color[HTML]{262626} 452.6}                       & {\color[HTML]{262626} 59.33}                        & {\color[HTML]{262626} 17.80}     & {\color[HTML]{262626} 98.73}     & {\color[HTML]{262626} 67.95}   \\
{\color[HTML]{262626} Sparse Local Attention}                              & {\color[HTML]{262626} 377.1}                       & {\color[HTML]{262626} 61.64}                        & {\color[HTML]{262626} 18.51}     & {\color[HTML]{262626} 98.81}     & {\color[HTML]{262626} 69.11}   \\ \hline
\end{tabular}%
}
\label{tab:moba}
\end{table}

\subsection{Comparison in T2V: End-to-End vs. Cascaded}

We also compare the performance of two different T2V paradigms: a large T2V model that directly generates 1080p videos (i.e., end-to-end), versus a large T2V model that first produces 512p latent representations followed by the SimpleGVR module to generate 1080p outputs (i.e., cascaded). As shown in Tab.~\ref{tab:t2v_paradigms}, the cascaded paradigm achieves significantly higher performance on quality metrics than the end-to-end paradigm. On other metrics that measure diverse aspects of videos (i.e., smoothness and consistency), the results under both paradigms are comparable.

\begin{table}[]
\centering
\caption{Quantitative comparison between two different T2V paradigms on AIGC100 dataset.}
\label{tab:t2v_paradigms}
\resizebox{1.0\textwidth}{!}{%
\begin{tabular}{c|c|ccc|cccccc}
\hline
{\color[HTML]{262626} }                         & {\color[HTML]{262626} }                        & \multicolumn{3}{c|}{{\color[HTML]{262626} DOVER}}                                                                     & \multicolumn{6}{c}{{\color[HTML]{262626} Vbench Metrics}}              \\ \cline{3-11} 
\multirow{-2}{*}{{\color[HTML]{262626} Method}} & \multirow{-2}{*}{{\color[HTML]{262626} MUSIQ}} & {\color[HTML]{262626} Technical}      & {\color[HTML]{262626} Aesthetic}      & {\color[HTML]{262626} Overall}        & {\color[HTML]{262626} \begin{tabular}[c]{@{}c@{}}Background\\ Consistency\end{tabular}} & {\color[HTML]{262626} \begin{tabular}[c]{@{}c@{}}Subject \\ Consistency\end{tabular}} & {\color[HTML]{262626} \begin{tabular}[c]{@{}c@{}}Aesthetic\\ Quality\end{tabular}} & {\color[HTML]{262626} \begin{tabular}[c]{@{}c@{}}Imaging\\ Quality\end{tabular}} & {\color[HTML]{262626} \begin{tabular}[c]{@{}c@{}}Motion\\ Smoothness\end{tabular}} & {\color[HTML]{262626} \begin{tabular}[c]{@{}c@{}}Average \\ Score\end{tabular}} \\ \hline
End-to-End                                      &   56.77                                       & 18.82                                 & 97.27                                 & 62.32                                 & {\color[HTML]{262626} 96.04}                                                            & {\color[HTML]{262626} 95.16}                                                          & {\color[HTML]{262626} 63.45}                                                       & {\color[HTML]{262626} 67.69}                                                     & {\color[HTML]{262626} 98.89}                                                       & {\color[HTML]{262626} 84.25}                                                    \\
Cascaded                                        & {\color[HTML]{262626} \textbf{62.35}}          & {\color[HTML]{262626} \textbf{20.44}} & {\color[HTML]{262626} \textbf{98.88}} & {\color[HTML]{262626} \textbf{71.34}} & {\color[HTML]{262626} 95.35}                                                            & {\color[HTML]{262626} 94.32}                                                          & {\color[HTML]{262626} 62.84}                                                       & {\color[HTML]{262626} \textbf{71.91}}                                                     & {\color[HTML]{262626} 98.74}                                                       & {\color[HTML]{262626} \textbf{84.63}}                                           \\ \hline
\end{tabular}
}
\end{table}

\section{Conclusions}

In this work, we propose SimpleGVR, a cascaded latent video super-resolution model designed to efficiently enhance the output of the Large T2V model into high-resolution video latent representations. To better align SimpleGVR with the base generator, we introduce two degradation strategies for synthesizing training pairs. We also investigate the effects of noise augmentation and timestep sampling to promote more accurate detail generation. In addition, to address the high computational cost of global full-attention, we propose two efficient design adaptations. Experimental results demonstrate the superiority of our framework, providing a strong baseline for future advancements in efficient cascaded VSR systems.

\bibliographystyle{unsrt}
\bibliography{ref}

\end{document}